\documentclass{ifacconf}

\usepackage{graphicx}      % include this line if your document contains figures
\usepackage{natbib}        % required for bibliography

\usepackage{siunitx}

\let\ifacconfcaptionwidth\captionwidth
\usepackage[caption=false]{subfig}
\let\captionwidth\ifacconfcaptionwidth

\usepackage{dblfloatfix}    % To enable figures at the bottom of page
\usepackage{booktabs}
\usepackage{multirow}
\usepackage{amsmath}
\begin{document}
% For arxiv file
\thispagestyle{empty}
\newpage
\onecolumn
\begin{center}
This paper has been accepted for publication in 2020 International Federation of Automatic Control (IFAC).
\vspace{0.75cm}\\
DOI: \\ % \href{https://doi.org/10.1109/ICRA.2019.8793718}{\textcolor{blue}{10.1109/ICRA.2019.8793718}}\\
IEEE Xplore: \\% \href{https://ieeexplore.ieee.org/document/8793718}{\textcolor{blue}{https://ieeexplore.ieee.org/document/8793718}}
\vspace{1.25cm}
\end{center}
©2020 the authors under a Creative Commons Licence CC-BY-NC-ND. Personal use of this material is permitted. Permission from IEEE must be obtained for all other uses, in any current or future media, including reprinting/republishing this material for advertising or promotional purposes, creating new collective works, for resale or redistribution to servers or lists, or reuse of any copyrighted component of this work in other works.
\twocolumn

\thispagestyle{empty}
\pagestyle{empty}

\begin{frontmatter}

\title{Deep Learning based Segmentation of Fish in Noisy Forward Looking MBES Images} 

%\thanks[footnoteinfo]{Sponsor and financial support acknowledgment
%goes here. Paper titles should be written in uppercase and lowercase
%letters, not all uppercase.}

\author[First]{Jesper Haahr Christensen}
\hspace{-0.25cm}\author[Second]{ }
%\author[Second]{Max Abildgaard} 
\author[Second]{Lars Valdemar Mogensen}
\author[First]{Ole Ravn}

\address[First]{Electrical Engineering Department, 
   Technical University of Denmark, 2800 Kgs. Lyngby, Denmark, \\ e-mail: \{jehchr,or\}@elektro.dtu.dk}
\address[Second]{ATLAS MARIDAN, 2960 Rungsted Kyst, Denmark, \\ e-mail: \{jhc,lvm\}@atlasmaridan.com}

\begin{abstract}                % Abstract of not more than 250 words.
In this work, we investigate a Deep Learning (DL) approach to fish segmentation in a small dataset of noisy low-resolution images generated by a forward-looking multibeam echosounder (MBES). 
We build on recent advances in DL and Convolutional Neural Networks (CNNs) for semantic segmentation and demonstrate an end-to-end approach for a fish/non-fish probability prediction for all range-azimuth positions projected by an imaging sonar. 
We use self-collected datasets from the Danish Sound and the Faroe Islands to train and test our model and present techniques to obtain satisfying performance and generalization even with a low-volume dataset.
We show that our model proves the desired performance and has learned to harness the importance of semantic context and take this into account to separate noise and non-targets from real targets.
Furthermore, we present techniques to deploy models on low-cost embedded platforms to obtain higher performance fit for edge environments -- where compute and power are  restricted by size/cost -- for testing and prototyping. 
\end{abstract}


\end{frontmatter}
%===============================================================================

\setcounter{figure}{1}
\begin{figure*}[b!]
    \centering
    \subfloat[]{
        \includegraphics[width=0.32\textwidth]{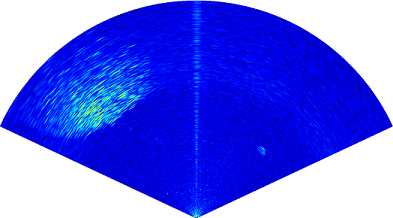}
    }
    \subfloat[]{
        \includegraphics[width=0.32\textwidth]{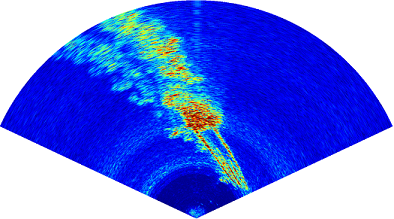}
    }
    \subfloat[]{
        \includegraphics[width=0.32\textwidth]{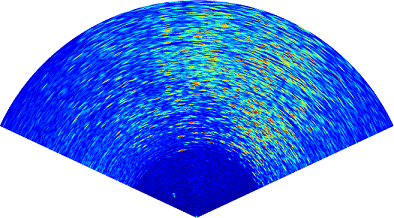}
    }
    \caption{Training set MBES data samples: (a) school of Herring, (b) surface vessel, (c) school of Herring.}
\label{fig:data:mbes}
\end{figure*}

\section{Introduction}
\label{sec:intro}
%Multibeam echo sounders (MBES) are the preferred choice for underwater imaging as optical sensors are limited in range and highly affected by turbidity in water. Even compact sonars in the low-price range offer ranges up to \SI{100}{\meter} with reasonable range and angular resolution. These are typically operated at frequencies around \SI{750}{\kilo\hertz} and \SI{1.2}{\mega\hertz}. 

In many circumstances, the preferred choice of an imaging device is an optical sensor. This produces high quality, high-resolution images with plenty of information on detailed content, colors, shapes, and textures. However, in certain environments, optical sensors are not well suited. In underwater imaging, optics are highly affected by turbidity in water which reduces visibility. This is both in terms of obscuring line of sight with particles and other organic detritus (marine snow) and in terms of illumination, which is attenuated with the amount of turbidity and depth. 

As such, a multibeam echosounder (MBES) is often preferred for underwater imaging. In terms of imaging information, it hardly compares with optics; however, the robustness to underwater conditions and extended range while maintaining a somewhat high resolution of contours makes MBES ideal. Even compact sonars integrable on small autonomous underwater vehicles (AUVs) or remotely operated vehicles (ROVs) offer ranges up to more than \SI{100}{\meter} with range and angular resolution dependent on application, model, and price. Sonar images do, however, suffer from distortion, noise and contain low-level detail and visual information only and do as such still pose a challenging task when processing sonar data for vision purposes.

In certain marine applications such as subsea monitoring, inspection tasks, or fisheries, it may be useful to have a small easy-portable system capable of simultaneously collecting and utilizing sonar data for autonomy or meta-data purposes. By using semantic-based segmentation masks, fish instances can be identified and measured in size and numbers. Fish schools can be identified at range, and absolute position can automatically be mapped, or relative position can feed into a navigation system. For databases with meta-data generated by such a model, queries can be made to specific situations and thus save hours and days of labor-intensive manual work of reviewing and labeling sonar data in extensive datasets.

The architectural choices for our segmentation model leverage recent advances in Deep Learning (DL) and Convolutional Neural Networks (CNNs) within segmentation for medical imaging~(\cite{u-net}) and scene understanding~(\cite{FCN,segnet,deeplab}). Our proposed model allows for end-to-end training and processing from input sonar image to binary output mask. 

As this work is highly application-oriented, we further demonstrate how we prepare our model for deployment in an edge environment, i.e., on-board online processing right where the data is created. Here size, power, and cost often limit the options for compute. We present tools to accomplish this and report performance on two typically used low-cost embedded platforms. 

%To obtain the better model and accuracy, two training strategies are explored; one seeks to first pre-train the network on a large publicly available segmentation dataset and then fine-tune the model towards the MBES data. The second strategy investigates the result achieved by training directly on the low-volume MBES data. 

The contributions in this work comprise a novel application proposal to obtain a target/non-target probability prediction for all range-azimuth positions as projected by a MBES on a pixel-based coordinate space using DL on the edge. Along with model architecture, techniques for training and obtaining high accuracy models with low data volumes are demonstrated. 
%The main contributions in this work are: 1) a novel approach to obtain a target/non-target probability prediction for all range-angular positions as projected by a MBES on a pixel-based coordinate space using deep learning. 2) a demonstration of techniques to obtain high accuracy models with low data volumes. 3) an evaluation of the robustness and generalization of the model to verify the models ability to learn from spatial context of fish or non-fish targets in sonar images. 
We also note that the proposed work does not limit itself within the application of fish segmentation but could potentially be expanded to any target or multi-class segmentation models. 

The remainder of this paper is organized as follows. Section~\ref{sec:related} provides a short literature survey of related work. Section~\ref{sec:data} describes the datasets and the collection of these. Section~\ref{sec:method} describes our method for fish segmentation. Section~\ref{sec:experiments} provides details of experiments. Section~\ref{sec:edge} demonstrates deployment and performance on the edge. Finally, in Section~\ref{sec:conclusion}, we present our conclusion.

\section{Related Work}
\label{sec:related}

\noindent\textbf{Deep Learning} \quad 
In line with CNNs revolutionizing the area of visual perception above water, increasing efforts are being made to apply such methods for underwater optical and acoustical sensing as well. One key challenge for applying such methods in the maritime domain is data availability. For autonomous driving and related fields, large annotated datasets containing millions of images are openly available, such as ImageNet~(\cite{imagenet}), Kitti~(\cite{kitti}), BDD100K~(\cite{BDD100K}) and Cityscapes~(\cite{Cityscapes}) to name a few. Sadly this is not yet the case in our field and thus in the case of a small dataset (\textless 5.000 samples) techniques such as data augmentation~(\cite{DataAug}) and transfer learning~(\cite{TransferLearning}) may be used. In \cite{FishObjectDetection}, an object detection model, using both before mentioned techniques, has been trained to detect, localize and classify fish and fish species in optical images. \cite{SonarCnnBestPractice,SonarObjectClassTransfer} applies transfer learning to perform image classification on specific cropped regions of sonar images. Similarly, \cite{seaturtle} train an object detection model to predict the presence and location of sea turtles in sonar images. 
\\

\noindent\textbf{Sonar Image Segmentation} \quad 
In \cite{RecCorals}, a na\"ive form of segmentation is achieved on high-resolution synthetic aperture sonar images for mapping corals in the data. This is carried out by dividing sonar trajectories into smaller segments and then perform a per-segment classification and reassemble the map to get a pseudo segmentation of the traversed area.~\cite{dos_Santos_2017} obtain a segmentation of sonar images by first identifying blobs or regions of interest in the image using traditional image processing techniques. Second, detected blobs are classified using linear machine learning models such as support vector machines, decision trees, and K-nearest neighbors. 

%To the best of our knowledge, no prior work in end-to-end deep learning based segmentation on MBES imaging has been published.

\section{Data}
\label{sec:data}
We introduce two separate datasets for training and testing our model. The data is collected using the same sensor but with two different vehicles at different locations and fish species. Although the visual appearance of the two datasets is similar, completely separated datasets for training and testing are a prerequisite in order to validate the generalization of the model rather than only testing the fit to the training data. 

\subsection{Training data details} \label{sec:data:train}
Our training set is collected autonomously in the Danish Sound off the coast of our offices in Rungsted (DK). This is carried out using our man-portable AUV shown in Fig.~\ref{fig:AUV}. The MBES data is collected using a forward-looking BluePrint Oculus m750d multibeam imaging sonar. Data is recorded with a range interval of $\left[0,20\right]\SI{}{\meter}$, an aperture of $\SI{130}{\degree}$ horizontal and $\SI{20}{\degree}$ vertical at an operating frequency of $\SI{750}{\kilo\hertz}$. 

\setcounter{figure}{0}
\begin{figure}
    \centering
    \includegraphics[width=1\linewidth]{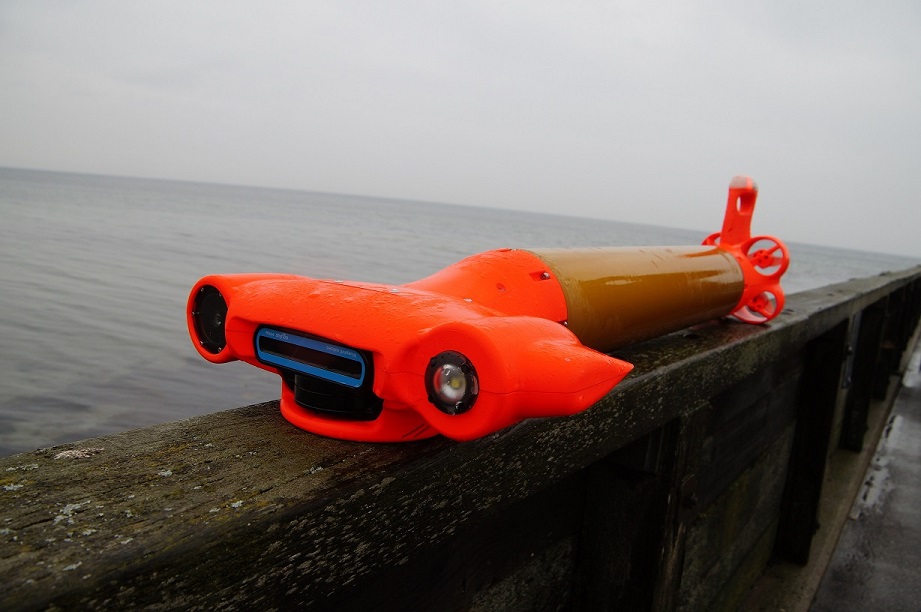}  
    \caption{Photo of the ATLAS MARIDAN man-portable AUV used for collecting training data.}
    \label{fig:AUV}
\end{figure}

\setcounter{figure}{3}
\begin{figure*}[b]
    \centering
	\includegraphics[width=0.9\textwidth]{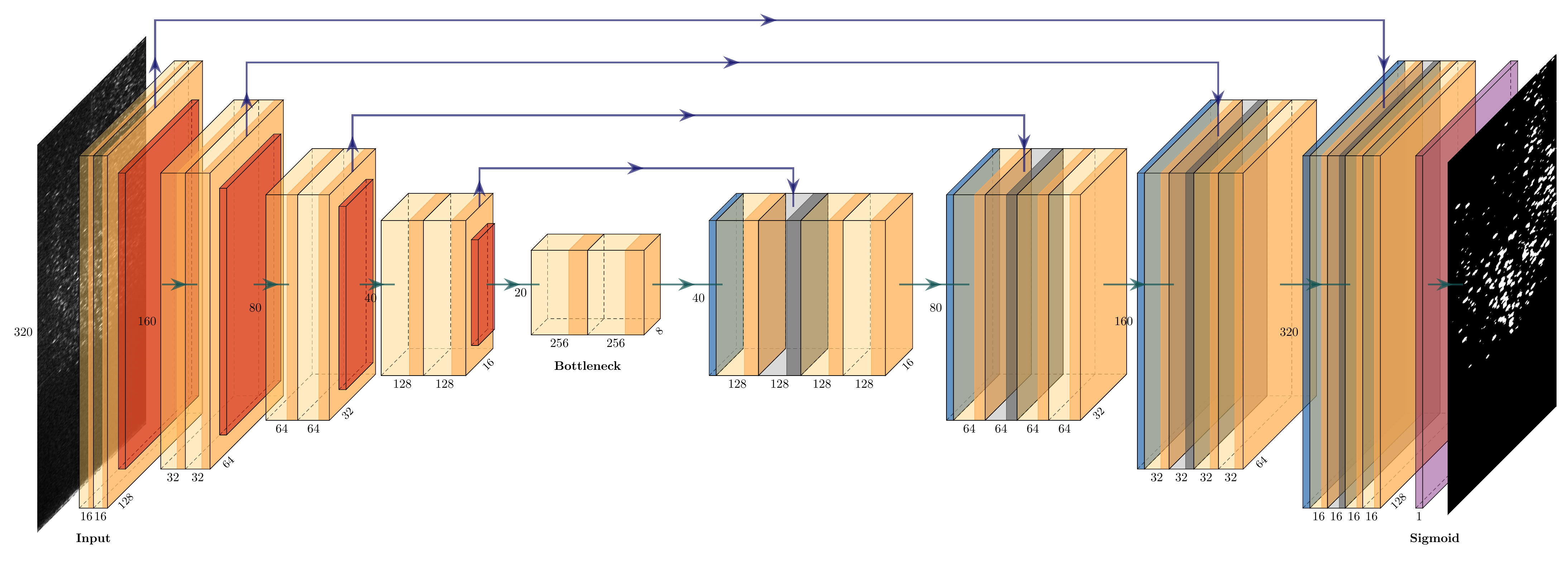}
    \caption{The encoder-decoder style network architecture for fish segmentation in sonar images. Input/output resolution is $320\times128$ and lowest resolution (in the bottleneck) is $20\times8$.}
\label{fig:architecture}
\end{figure*}

The labeled training set contains in total 50 images, which include more than $\SI{5000}{}$ targets (Herring) along with non-fish data such as surface reflections, surface vessels, and bottom returns. The dataset is annotated using the LabelBox annotation tool. Three samples from the training set are shown in Fig.~\ref{fig:data:mbes}.
Far from all fish in every image have been annotated due to the extensive process of manually drawing segmentation masks.

\subsection{Test data details} \label{sec:data:test}
Our test set is collected in the fjords of the Faroe Islands. This is carried out using our ROV sensor test rig shown in Fig.~\ref{fig:data:faroe}(a). The sensor and settings are similar to those described in Section~\ref{sec:data:train}. The recordings differ with varying range settings, which scales the visual appearance. The targets of the Faroe Islands are of the fish specie Wittling, which slightly differs from Herring in size, numbers, and behavior. The setup and a sample are shown in Fig.~\ref{fig:data:faroe}.

\setcounter{figure}{2}
\begin{figure}
    \centering
    \subfloat[]{
        \includegraphics[width=0.85\linewidth]{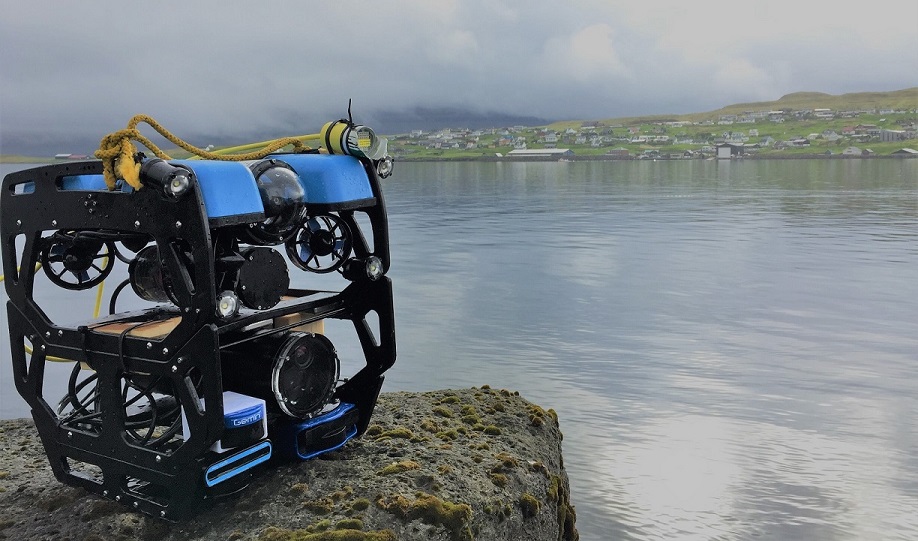}
    }
    
    \subfloat[]{
        \includegraphics[width=0.85\linewidth]{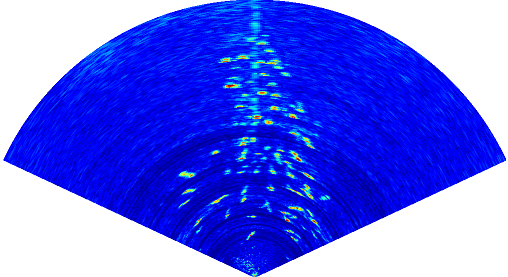}
    }
    
    \subfloat[]{
        \includegraphics[width=0.85\linewidth]{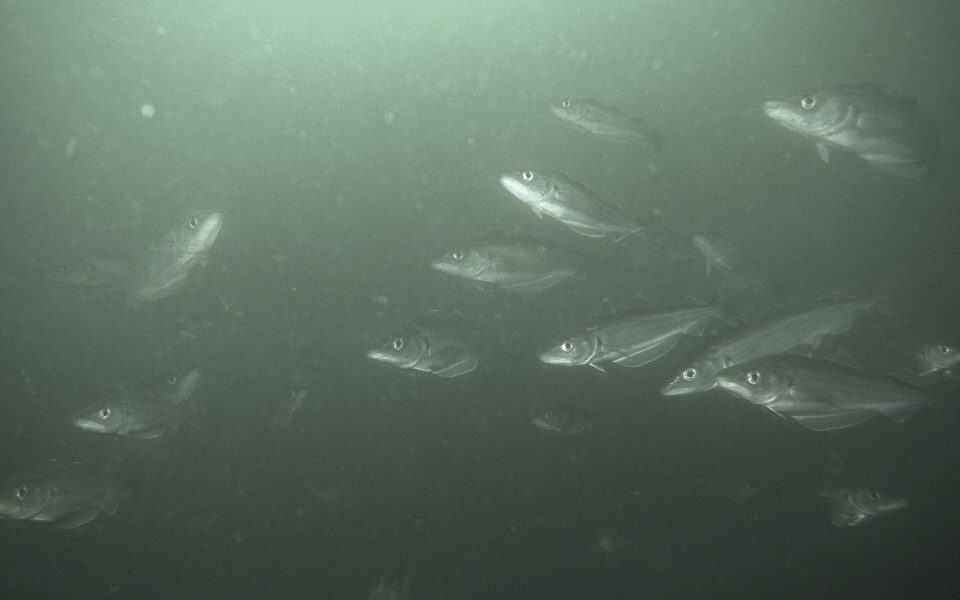}
    }
    \caption{Test data collection at the Faroe Islands: (a) Our ROV test rig for sensors, (b) Sonar image sample, (c) Optical image sample.}
\label{fig:data:faroe}
\end{figure}

\section{Method}
\label{sec:method}

Our network architecture is shown in Fig.~\ref{fig:architecture} and builds upon typical convolutional encoder-decoder style networks. Here we embed our input in a latent feature space and then seek to reconstruct the input as semantic classes. We employ skip connections to fully recover the fine-grained spatial information lost in pooling or down-sampling layers. This is as proposed by \cite{FCN} and used in nearly all following work on segmentation.
Table~\ref{tab:architecture} and the following describes the architecture and dimensions in more detail. 

Input dimensions are $1\times 320 \times 128$ corresponding to the 1-channel MBES data at half resolution represented in a Cartesian coordinate space. As most targets are not more than a few pixels in height, this is the borderline resolution without losing too much information.

Each {\tt Conv Layer} combines a sequence of a convolution, batch normalization, dropout, convolution, and batch normalization. This seeks to obtain an efficient and regularized training progress. The ReLU non-linearity is used as activation function for each {\tt Conv Layer}.
Each convolution is furthermore padded in order to keep input and output dimensions consistent.

{\tt Up-sample} combine transpose convolutions and a fusion with the corresponding feature map in the encoding path. 

Finally, a {\tt Sigmoid Layer} is applied. This combines a $1\times 1$ convolution to down-sample the channel dimension to a single output channel and a sigmoid activation to reach the final output as a probability for each pixel. The sigmoid output is then threshold at $0.5$ to produce either 1 or 0 as the final output for the two segmentation classes; fish and non-fish.

\addtolength{\tabcolsep}{-1pt} 
\begin{table}[]
\caption{Architecture of fish segmentation network.}
\begin{tabular}{lllcc}
\multicolumn{1}{l|}{}                                                                         & \multicolumn{1}{l|}{Name}          & \multicolumn{1}{l|}{} & \multicolumn{1}{c|}{\begin{tabular}[c]{@{}c@{}}Feat maps\\ (input)\end{tabular}} & {\begin{tabular}[c]{@{}c@{}}Feat maps\\ (output)\end{tabular}} \\ \hline
\multicolumn{1}{l|}{\multirow{8}{*}{\begin{tabular}[c]{@{}l@{}}Encoding\\ path\end{tabular}}} & \multicolumn{1}{l|}{Conv Layer 1}  & \multicolumn{1}{l|}{} & \multicolumn{1}{c|}{$1 \times 320\times 128$}                                    & $16 \times 320\times 128$                                    \\
\multicolumn{1}{l|}{}                                                                         & \multicolumn{1}{l|}{Max-pooling 1} & \multicolumn{1}{l|}{} & \multicolumn{1}{c|}{$16 \times 320\times 128$}                                   & $16 \times 160\times 64$                                    \\
\multicolumn{1}{l|}{}                                                                         & \multicolumn{1}{l|}{Conv Layer 2}  & \multicolumn{1}{l|}{} & \multicolumn{1}{c|}{$16 \times 160\times 64$}                                    & $32 \times 160\times 64$                                    \\
\multicolumn{1}{l|}{}                                                                         & \multicolumn{1}{l|}{Max-pooling 2} & \multicolumn{1}{l|}{} & \multicolumn{1}{c|}{$32 \times 160\times 64$}                                    & $32 \times 80 \times 32$                                    \\
\multicolumn{1}{l|}{}                                                                         & \multicolumn{1}{l|}{Conv Layer 3}  & \multicolumn{1}{l|}{} & \multicolumn{1}{c|}{$32 \times 80 \times 32$}                                    & $64 \times 80 \times 32$                                   \\
\multicolumn{1}{l|}{}                                                                         & \multicolumn{1}{l|}{Max-pooling 3} & \multicolumn{1}{l|}{} & \multicolumn{1}{c|}{$64 \times 80 \times 32$}                                    & $64 \times 40 \times 16$                                    \\
\multicolumn{1}{l|}{}                                                                         & \multicolumn{1}{l|}{Conv Layer 4}  & \multicolumn{1}{l|}{} & \multicolumn{1}{c|}{$64 \times 40 \times 16$}                                    & $128 \times 40 \times 16$                                    \\
\multicolumn{1}{l|}{}                                                                         & \multicolumn{1}{l|}{Max-pooling 4} & \multicolumn{1}{l|}{} & \multicolumn{1}{c|}{$128 \times 40 \times 16$}                                   & $128 \times 20 \times 8$                                    \\ \hline
\multicolumn{1}{l|}{}                                                                         & \multicolumn{1}{l|}{Bottleneck}    & \multicolumn{1}{l|}{} & \multicolumn{1}{c|}{$128 \times 20 \times 8$}                                    & $256 \times 20 \times 8$                                    \\ \hline
\multicolumn{1}{l|}{\multirow{9}{*}{\begin{tabular}[c]{@{}l@{}}Decoding\\ path\end{tabular}}} & \multicolumn{1}{l|}{Up-sample 1}   & \multicolumn{1}{l|}{} & \multicolumn{1}{c|}{$256 \times 20 \times 8$}                                    & $256 \times 40 \times 16$                                    \\
\multicolumn{1}{l|}{}                                                                         & \multicolumn{1}{l|}{Conv Layer 5}  & \multicolumn{1}{l|}{} & \multicolumn{1}{c|}{$256 \times 40 \times 16$}                                   & $128 \times 40 \times 16$                                    \\
\multicolumn{1}{l|}{}                                                                         & \multicolumn{1}{l|}{Up-sample 2}   & \multicolumn{1}{l|}{} & \multicolumn{1}{c|}{$128 \times 40 \times 16$}                                   & $128 \times 80 \times 32$                                   \\
\multicolumn{1}{l|}{}                                                                         & \multicolumn{1}{l|}{Conv Layer 6}  & \multicolumn{1}{l|}{} & \multicolumn{1}{c|}{$128 \times 80 \times 32$}                                   & $64 \times 80 \times 32$                                   \\
\multicolumn{1}{l|}{}                                                                         & \multicolumn{1}{l|}{Up-sample 3}   & \multicolumn{1}{l|}{} & \multicolumn{1}{c|}{$64 \times 80 \times 32$}                                    & $64 \times 160 \times 64$                                  \\
\multicolumn{1}{l|}{}                                                                         & \multicolumn{1}{l|}{Conv Layer 7}  & \multicolumn{1}{l|}{} & \multicolumn{1}{c|}{$64 \times 160 \times 64$}                                   & $32 \times 160 \times 64$                                   \\
\multicolumn{1}{l|}{}                                                                         & \multicolumn{1}{l|}{Up-sample 4}   & \multicolumn{1}{l|}{} & \multicolumn{1}{c|}{$32 \times 160 \times 64$}                                   & $32 \times 320 \times 128$                                   \\
\multicolumn{1}{l|}{}                                                                         & \multicolumn{1}{l|}{Conv Layer 8}  & \multicolumn{1}{l|}{} & \multicolumn{1}{c|}{$32 \times 320 \times 128$}                                  & $16 \times 320 \times 128$                                   \\
\multicolumn{1}{l|}{}                                                                         & \multicolumn{1}{l|}{Sigmoid Layer} & \multicolumn{1}{l|}{} & \multicolumn{1}{c|}{$16 \times 320 \times 128$}                                  & $1 \times 320 \times 128$   
\end{tabular}
\label{tab:architecture}
\end{table}
\addtolength{\tabcolsep}{1pt} 

Since we have two classes only (fish, non-fish), we can learn the mapping from input sonar images to output masks using binary cross-entropy loss:
\begin{align}
\begin{split}
    \mathcal{L}_{\textrm{BCE}}(M) &= -\sum^{C=2}_{i=1}y_i \log{\left[M(x_i)\right]} \\
    &=-y_1 \log{\left[M(x_1)\right]} - (1-y_1)\log{\left[1-M(x_1)\right]}
\end{split}
\end{align}
where $M$ is our model, $C$ is our two classes, $i$ represents each class in $C$, $y$ is the ground truth pixel value, and $x$ is the input data. 

Our objective therefore is:
\begin{align}
    \min_M \mathcal{L}_{\textrm{BCE}}(M).
\end{align}

\section{Experiments}
\label{sec:experiments}
%In this section a detailed description on how the models are trained is given. Then experiments on the trained models are carried to assess the performance and robustness of these.

The model is trained using 
%an cloud Amazon Web Services instance utilizing 
one NVIDIA Tesla K80 GPU. As the architecture is fairly light-weight in terms of parameters, time to reach convergence is less than \SI{1}{\hour}. The model is implemented in the TensorFlow DL framework using the Keras frontend for Python. 
During training, we use a batch size of 4 and all weights are initialized using the He normal initializer~(\cite{He_activation}). All ReLUs are leaky with slope $0.2$.
%Our models seek to minimize the binary crossentropy loss as
%\begin{align}
%    \mathcal{L}_{\textrm{BCE}} = -\sum^{C=2}_{i=1}y \log{\hat{y}}=-y \log{\hat{y}} - (1-y)\log{1-\hat{y}}
%\end{align}
%where $C$ is the two classes, $y$ is the ground truth pixel value and $\hat{y}$ is the predicted pixel value. 
The loss function is minimized using the recently introduced RAdam solver~(\cite{liu2019variance}) with initial learning rate $\eta = 0.5\times10^{-4}$ and parameters $\beta_1$ and $\beta_2$ set to $0.9$ and $0.999$ respectively. 

With these hyperparameters, the model converges within 100 epochs to a validation loss and an accuracy of \SI{0.059}{} and \SI{97}{\percent}, respectively. Additionally, the precision, recall, and F1-score for the fish-class 
(the content we are interested in) 
are \SI{69}{\percent}, \SI{81}{\percent}, and \SI{75}{\percent}, respectively. %in about \SI{20000}{} training steps. 
The final model size is 22~MB.

To overcome issues with low data volume, the data is heavily augmented at training time. This is done by randomly applying horizontal and vertical flipping, rotations in range $\left[\SI{-20}{},\SI{20}{}\right]\SI{}{\degree}$, width and height shifts in range $\left[\SI{-20}{},\SI{20}{}\right]\SI{}{\percent}$, and random cropping.

%The training and validation accuracy and loss for all three training passes are shown in Fig.~\ref{fig:training_plots}. On the validation dataset, the model trained directly on the MBES data achieves \SI{97.96}{\percent} pixel accuracy and a loss of \SI{0.059}{} in about \SI{2000}{} training steps. 
%The model first pre-trained on the nuclei dataset and then fine-tuned on the MBES dataset achieves \SI{97.79}{\percent} pixel accuracy and a loss of \SI{0.057}{} in about \SI{950}{} training steps excluding the pre-training. 

We test our model on the dataset obtained in the fjords of the Faroe Islands. This allows for a qualitative assessment only, as we have no ground-truth data due to the labor-intensive work going into creating this. It will, however, indicate the performance, robustness, and generalization achieved by the model. Fig.~\ref{fig:val_grid} shows nine samples of sonar images and the predicted output mask from the trained model. The samples clearly show that the model has learned to discriminate between fish and non-fish and thus predicts noise, surface reflections, bottom returns, and the like as non-fish. 

As explained in Section~\ref{sec:data:test}, the targets in the test data are no longer Herring, but a bigger fish specie named Wittling. The visual (sonar image) appearance of these fish is similar at an instance-level. However, the schooling behavior is unlike Herring, and for much of the obtained data Wittlings are notably more scattered and at a closer range than Herring in the training data. 

\setcounter{figure}{5}
\begin{figure*}[t!]
    \centering
    \subfloat{
        \includegraphics[width=0.98\textwidth]{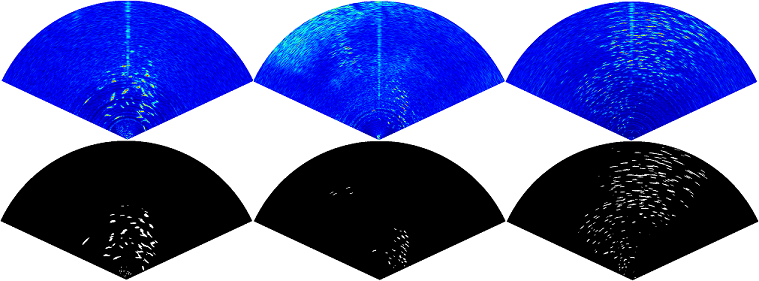}
    }
    \vspace{10pt}
    \subfloat{
        \includegraphics[width=0.98\textwidth]{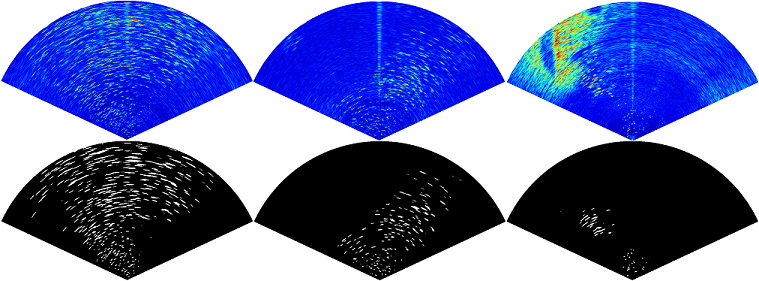}
    }
    \vspace{10pt}
    \subfloat{
        \includegraphics[width=0.98\textwidth]{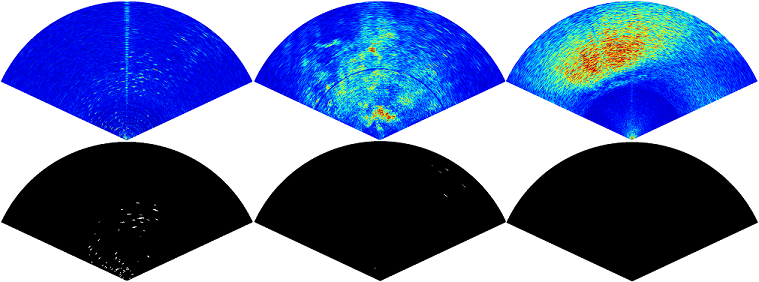}
    }
    \caption{Test images from the Faroe Islands dataset. Sonar image is on top, predicted mask on bottom.}
\label{fig:val_grid}
\end{figure*}

\section{Edge implementation} \label{sec:edge}
Operating at the edge refers to a setting where data is processed at the location of the sensor. Typically for small submersible systems, the budget for cost, compute, power, and size for processing platforms is restricted. Two popular choices are a Raspberry Pi and an NVIDIA Jetson Nano. Table~\ref{tab:edge} summarizes performance measures on post-training optimized models on a single CPU core on a Raspberry Pi and on an NVIDIA Jetson Nano using its embedded GPU. Details are presented in the following.

As this work is highly application-oriented and part of ongoing work at ATLAS MARIDAN, the success of a model relies not only on results achieved of the model but also on its ability to be running in an embedded environment on the edge. Here we demonstrate how models quick and straightforward can be prepared for such environments while still maintaining a Python environment for convenient prototyping during inference. We report performance on two common and popular embedded boards without requirements for external accelerators such as the Intel Movidius Neural Compute Stick%\footnote{\url{https://software.intel.com/en-us/neural-compute-stick}}
or the Google Coral USB Accelerator%\footnote{\url{https://coral.withgoogle.com/products/accelerator}}
. 

\addtolength{\tabcolsep}{-0.5pt} 
\begin{table}[]
\caption{Edge performance.}
\centering
\begin{tabular}{@{}lcccc@{}}
\toprule
               & \begin{tabular}[c]{@{}c@{}}Default\\ (non-optimized)\end{tabular} & \begin{tabular}[c]{@{}c@{}}TFLite\\ (RPi3)\end{tabular} & \begin{tabular}[c]{@{}c@{}}TFLite\\ (RPi4)\end{tabular} & \begin{tabular}[c]{@{}c@{}}TensorRT\\ (Nano)\end{tabular} \\ \midrule
FPS            & 0.37 / 0.76 / 11                                                  & 0.45                                                    & 1.43                                                    & 33                                                        \\
Speedup        & -                                                                 & 1.2x                                                    & 1.88x                                                   & 3x                                                        \\
Model size     & 22.5 MB                                                           & 2.4 MB                                                  & 2.4 MB                                                  & 11.5 MB                                                   \\
Size reduction & -                                                                 & 9.375x                                                  & 9.375x                                                  & 1.96x                                                     \\ \bottomrule
\end{tabular}
\label{tab:edge}
\end{table}
\addtolength{\tabcolsep}{1pt} 

\noindent\textbf{Raspberry Pi 3/4} \quad %\subsubsection{Raspberry Pi 3/4}
Using the TensorFlow framework, we first save the model as a static inference graph and second quantize the weights of the graph to 8-bit precision rather than 64-bit. %In general this is without much loss of accuracy. 
The quantization is performed using the TensorFlow Lite framework for on-device inference. The post-quantization model size is 2.4~MB and does inference at 1.3 FPS using a single core only (out of 4). This is a more than $\times 9$ reduction in model size and a nearly $\times 2$ speedup. 

\noindent\textbf{NVIDIA Jetson Nano} \quad%\subsubsection{NVIDIA Jetson Nano}
Using the high-performance inference platform TensorRT from NVIDIA, we can use in-build support in TensorFlow to convert our model to a post-training quantized 16-bit TensorRT graph. As above, we first save our model as a static inference graph and then convert the graph to a quantized TensorRT graph. 

The TensorRT model size is 11.5~MB and does inference at 33 FPS using the embedded GPU on the Jetson Nano.  This is a $\times 2$ reduction in model size and a $\times 3$ speedup. 

\section{Conclusion}
\label{sec:conclusion}
                                  
%By visually inspecting the results and analyzing the training loss/accuracy graphs in Fig.~\ref{fig:training_plots}, it is found that the accuracy is about \SI{98}{\percent} and close to identical for the two models. The one thing gained from pre-training is a better initialization of the weights in the network which causes faster convergence of the model. However, in this particular case this yields a marginally lower accuracy than training directly on the MBES dataset.

Our model converges on the validation dataset with \SI{97}{\percent} accuracy. The precision, recall, and F1-score for fish targets are \SI{69}{\percent}, \SI{81}{\percent}, and \SI{75}{\percent}, respectively. Additionally, the results presented from the test dataset in Fig.~\ref{fig:val_grid} clearly show that our model has learned the desired functionality of segmenting fish-like targets from noise, surface reflections, and other non-fish objects in sonar images. Some targets remain to be recognized, and the model may be affected by the low data volume or bias caused by incomplete annotations. Possible improvements could be achieved by retraining with new data processed by the model in a semi-supervised manner. %However, it is clearly shown that the model has learned to take the semantic context into account and performs generally very good in separating noise and non-targets from actual targets even though instance shapes and pixel intensity may be similar. 

From the performance summary in Table~\ref{tab:edge}, it is shown that the model is capable of performing inference at suitable processing times on low-cost embedded devices. We obtain 33 FPS on a Jetson Nano and more than 1 FPS using only one core (out of 4), and nearly no memory on a Raspberry Pi 4. 1 FPS is deemed sufficient for most monitoring and long-range tracking applications at ATLAS MARIDAN.

%\begin{ack}
%This work was supported by ATLAS MARIDAN ApS and the thyssenkrupp funded start-up oXeanpedia. Thanks to colleagues at ATLAS MARIDAN who have contributed to the work in terms of AUV system development and data collection.
%\end{ack}

\bibliography{ifacconf}             % bib file to produce the bibliography
                                                     % with bibtex (preferred)

%\appendix
%\section{A summary of Latin grammar}    % Each appendix must have a short title.
%\section{Some Latin vocabulary}              % Sections and subsections are supported  
                                                                         % in the appendices.
\end{document}